\documentclass[10pt,twocolumn,letterpaper]{article}

\usepackage{cvpr}              
\renewcommand{\abstract}{%
  \centerline{\large\bf Abstract}%
  \vspace*{4pt}
  \noindent\it\ignorespaces
}

\newcommand{\PAR}[1]{\vspace{-0.2eM}\vskip4pt \noindent{\bf #1}}
\usepackage{multirow} 
\usepackage{makecell}
\usepackage[table, dvipsnames, svgnames, x11names]{xcolor}

\usepackage[dvipsnames]{xcolor}

\usepackage{adjustbox}
\usepackage{tabularx, booktabs, array}

\usepackage[mathlines]{lineno}
\usepackage{nccmath}

\usepackage{graphicx}

\newcommand{\truckdriveemoji}{%
  \raisebox{-0.22em}{\includegraphics[height=1.3em,trim=0 50 0 0,clip]{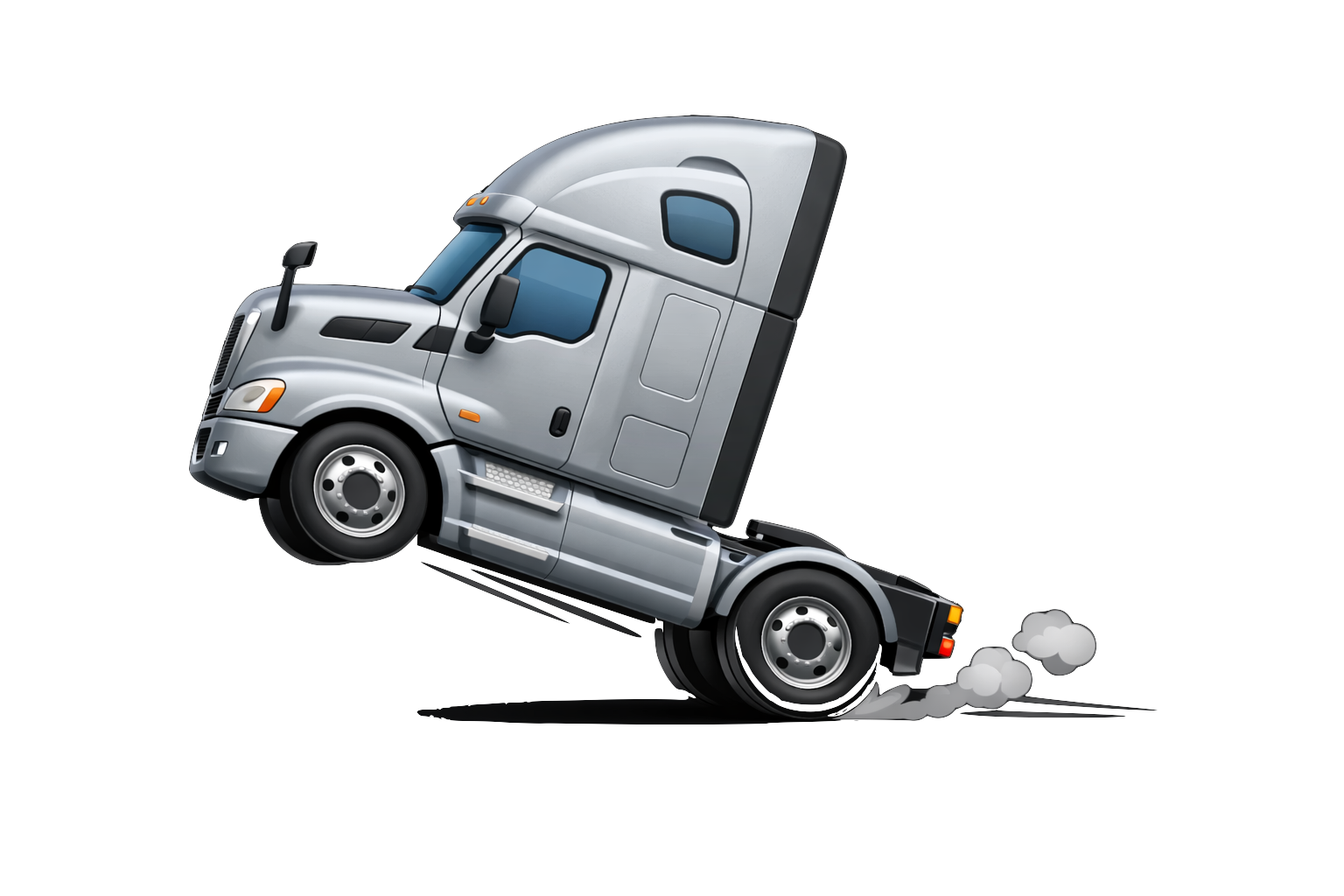}}}

\definecolor{cvprblue}{rgb}{0.21,0.49,0.74}
\usepackage[pagebackref,breaklinks,colorlinks,allcolors=cvprblue]{hyperref}

\def\paperID{\#\#\#\#} 
\def\confName{CVPR}
\def\confYear{2026}

\title{\truckdriveemoji TruckDrive: Long-Range Autonomous Highway Driving Dataset}

\author{
\begin{tabular}[t]{c}
Filippo Ghilotti$^{1}$ \hspace{1eM}
Edoardo Palladin$^{1}$ \hspace{1eM}
Samuel Brucker$^{1}$ \hspace{1eM}
Adam Sigal$^1$ \\
Mario Bijelic$^{1,2}$ \hspace{1eM} 
Felix Heide$^{1,2}$ 
\end{tabular}\\[0.2em]
\begin{tabular}[t]{c}
\small{$^1$Torc Robotics} \qquad \quad
\small{$^2$Princeton University} \qquad \quad
\end{tabular}
}

\makeatletter
\newcommand{\blfootnote}[1]{%
  \begingroup
  \renewcommand\thefootnote{}%
  \footnote{#1}%
  \addtocounter{footnote}{-1}%
  \endgroup
}
\makeatother

\begin{document}
\twocolumn[{
\maketitle
\thispagestyle{empty}
\pagestyle{empty}

\renewcommand\twocolumn[1][]{#1}
\vspace{-10mm}
\begin{center}
    \centering
    \captionsetup{type=figure}
    \includegraphics[width=.95\textwidth]{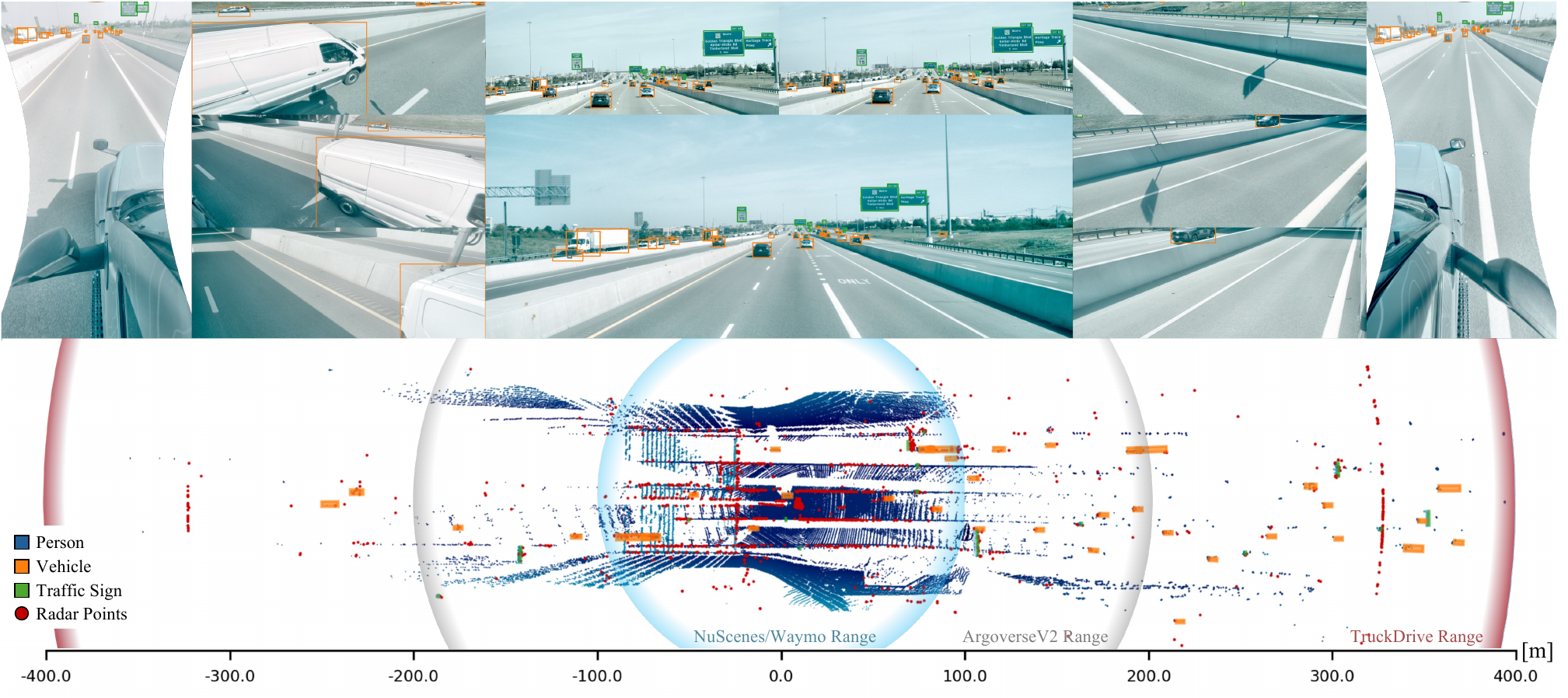}
    \vspace{-3mm}
    \captionof{figure}{\textbf{TruckDrive Dataset.} Autonomous vehicles, especially heavy trucks, require long planning horizons for safe driving in highway scenarios due to higher speed and longer breaking distances. This requires perception ranges well beyond $300$\,m, while the most common datasets are limited to $100$\,m \cite{caesar2020nuscenes, sun2020scalability}. We introduce the TruckDrive Dataset, a large scale multi-modal benchmark captured with a sensor setup tailored for long-range perception with LiDAR, radar and $3$D annotations up to $400$\,m and images and $2$D annotations up to $1000$\,m.}
    \label{fig:teaser}
\end{center}
}]

\blfootnote{All authors contributed equally to this work.} 

\begin{abstract}
Safe highway autonomy for heavy trucks remains an open and unsolved challenge: due to long braking distances, scene understanding of hundreds of meters is required for anticipatory planning and to allow safe braking margins. However, existing driving datasets primarily cover urban scenes, with perception effectively limited to short ranges of only up to 100 meters. To address this gap, we introduce TruckDrive, a highway-scale multimodal driving dataset, captured with a sensor suite purpose-built for long range sensing: seven long-range FMCW LiDARs measuring range and radial velocity, three high-resolution short-range LiDARs, eleven 8MP surround cameras with varying focal lengths and ten 4D FMCW radars. The dataset offers 475 thousands samples with 165 thousands densely annotated frames for driving perception benchmarking up to 1,000 meters for 2D detection and 400 meters for 3D detection, depth estimation, tracking, planning and end to end driving over 20 seconds sequences at highway speeds. We find that state-of-the-art autonomous driving models do not generalize to ranges beyond 150 meters, with drops between 31\% and 99\% in 3D perception tasks, exposing a systematic long-range gap that current architectures and training signals cannot close. Dataset download, devkit, and videos are available at: \href{https://light.princeton.edu/TruckDrive}{light.princeton.edu/TruckDrive}.
\end{abstract}
    
\section{Introduction}
\label{sec:intro}

Autonomous driving methods require scene understanding, robotic planning and control, either implicitly, in an end-to-end fashion, or in explicit modules, including perception~\cite{yin2021centerbased3dobjectdetection,fan2023fsdv2improvingfully,li2022bevformer, bai2022transfusionrobustlidarcamerafusion, liu2022bevfusion}, prediction of scene geometry \cite{piccinelli2025unidepthv2, guan2024neural} and of the future evolution of relevant agents \cite{girgis2022latent}, tracking of the environment \cite{ding20233dmotformer,yin2021centerbased3dobjectdetection, zhang2022mutr3d, wang2021immortal}, planning and control \cite{hu2023_uniad,jiang2023vad}.

In the last decade, the development of driving methods have been largely driven by learned models trained on driving datasets including KITTI~\cite{geiger2012we},  Cityscapes~\cite{Cordts2016Cityscapes}, nuScenes~\cite{caesar2020nuscenes}, Waymo~\cite{sun2020scalability} and Argoverse~\cite{Argoverse, Argoverse2}, which predominantly feature urban environments and therefore implicitly bias the development of the field toward short-range perception and low-speed driving. This bias is reflected in the annotation range, which typically extends only $70$–$100$\,m from the ego-vehicle.

For normal passenger cars driving in urban environments, short-range perception is sufficient as lower speeds convert the limited spatial range into enough temporal foresight to support the $5$–$10$ seconds planning horizons of modern prediction and planning stacks~\cite{Ettinger_2021_ICCV, Yadav_2024, 10588728, orbis2025}. For heavy-duty trucks operating at highway speeds, instead, the safety envelope is dominated by their high-inertia braking requirements. At $120$\,km/h, a fully-loaded truck requires over $150$–$200$\,m to stop, equivalent to $4.5$–$6$\,s of look-ahead perception. Therefore, the necessary braking budget is severely compromised by limited sensing horizons: an $80$\,m range provides only about $2.4$ seconds of foresight, and even $100$\,m yields merely $3.0$ seconds. This entire window is consumed by sensing and planning latencies, eroding the time required for safe braking actuation before the maneuver can even begin. This leaves a critically insufficient margin for the vehicle's deceleration and renders strategic planning, like merging or lane changes, unfeasible. 

\begin{figure}[t]
    \centering
       \includegraphics[trim=0.2cm 0.3cm 0cm 0cm, clip, width=\linewidth]{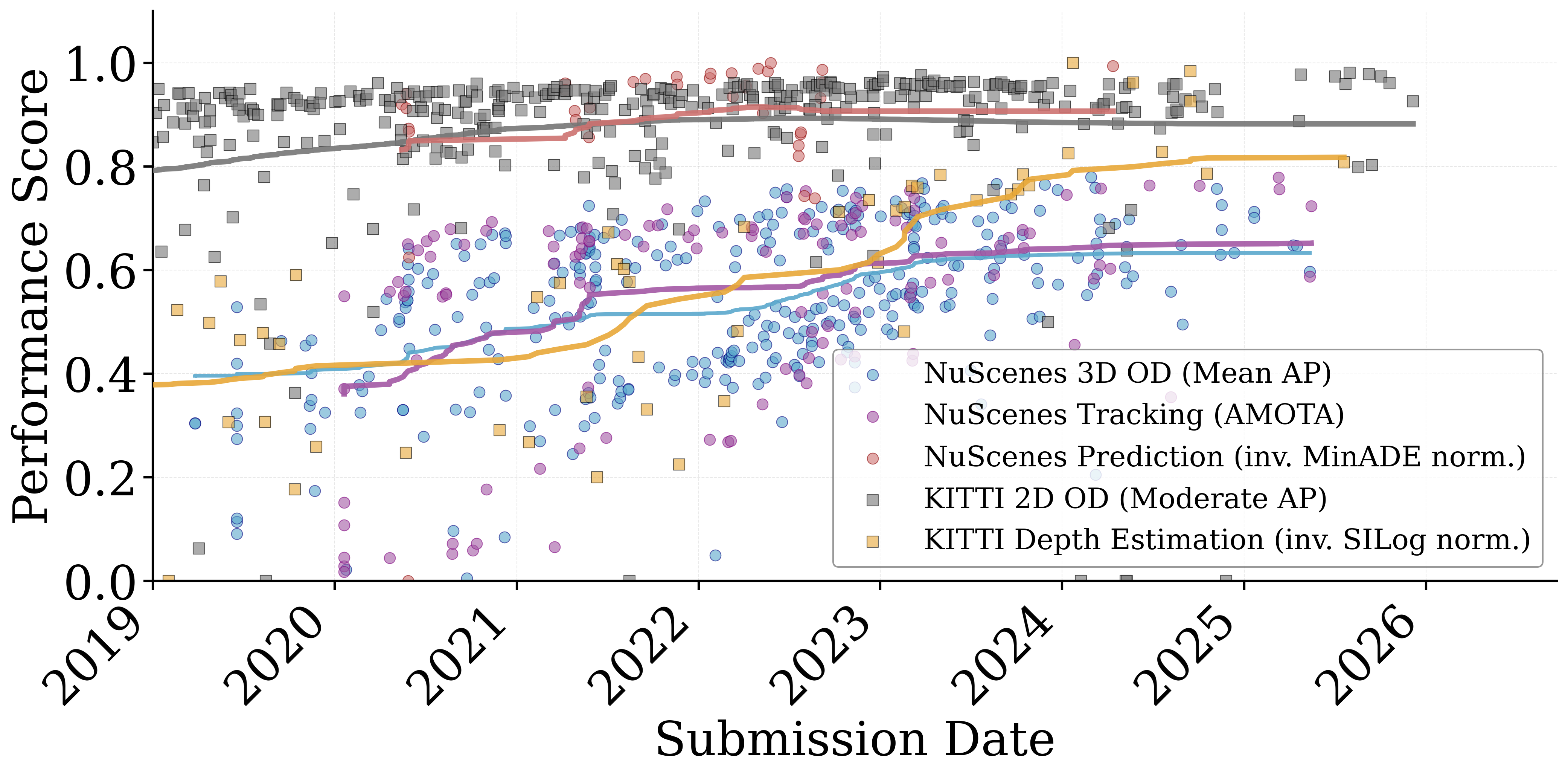}
       \vspace{-5mm}
    \caption{\textbf{Performance Saturation on Urban Datasets. }We plot the performance of 2D and 3D OD, Tracking, Prediction and Depth Estimation of NuScenes \cite{caesar2020nuscenes} and Kitti \cite{behley2020benchmarklidarbasedpanopticsegmentation} leader boards across the years and observe a saturation of these benchmarks. 
    }
    \label{fig:perf_saturation}
    \vspace{-5mm}
\end{figure}

Driving architectures for long-range perception and planning are non-trivial: Bird’s-Eye-View (BEV) and dense voxel representations scale quadratically with distance, leading to exponential growth in compute and memory~\cite{PalladinAndBruckerLRS4Fusion}. Concurrently, the signal-to-noise ratio of distant objects decreases sharply due to sensor resolution limits and atmospheric attenuation. Sparse \cite{xie2023sparsefusion} and range-aware methods~\cite{jiang2023far3dexpandinghorizonsurroundview} alleviate these issues but remain constrained by calibration drift, temporal uncertainty and sparsity of long-range supervision. Moreover, performance on short-range urban benchmarks has begun to saturate, with a decline in the number of submissions and a flattening in the performance gain (Figure~\ref{fig:perf_saturation}), with poor generalization capability beyond $100$ m~\cite{PalladinAndBruckerLRS4Fusion} of models designed around these priors.

To close this gap, we introduce \emph{TruckDrive}, the first large-scale dataset specifically designed for long-range, high-speed autonomous driving. TruckDrive, as presented in Figure \ref{fig:teaser}, extends the perception range by a factor of \emph{five} relative to urban benchmarks, compared in Table \ref{tab:dataset_comparison}, providing $2$D annotations up to $1{,}000$\,m and corresponding $3$D annotations up to $400$\,m, with $15$ to $25$ s temporal clips to support forecasting and end-to-end (E2E) learning. The dataset includes over $475$\,k multi-modal synchronized samples, among which $165$\,k manually labeled and $310$\,k unlabeled for self-supervised and unsupervised research.
Our sensor suite integrates high-resolution ($8$MP) short and long focal length cameras, wide-baseline stereo, short and long range $4$D LiDARs and $4$D radars, enabling comprehensive research in perception, prediction and planning.

We evaluate state-of-the-art driving methods for urban datasets in diverse tasks and observe drops between $31$ and $99$\% in $3$D perception tasks beyond $150$ m, confirming that they do not generalize to long-range regimes. This exposes a fundamental open challenge in current architectures and motivates new directions in efficient representation learning, sensor fusion and long-horizon reasoning.

We make the following contributions:

\begin{itemize}
\item We present a long-range, high-fidelity multi-modal driving dataset that combines high-resolution $8$MP cameras, large-baseline stereo, $4$D LiDARs and $4$D radars, enabling dense $3$D annotations up to $400$\,m and $2$D annotations up to 1\,km.
\item We provide large-scale data comprising $475$\,k samples, including $165$\,k labeled and $310$\,k unlabeled frames, with full raw sensor streams to support supervised, semi-supervised, and self-supervised research.
\item We establish a highway-scale benchmark for perception, prediction, planning and E2E driving tasks under high-speed, long-range conditions, finding several failure modes and scaling challenges in existing models.
\end{itemize}

\begin{table*}[ht!]
    \caption{\textbf{TruckDrive Benchmark Comparison.} Cross-dataset summary of sensors, synced samples and useful ranges. TruckDrive couples $7$ long range and $3$ short range LiDARs with $10$ automotive radars, $9$ wide/medium field of view cameras and $1-3$ long-focal-length wide-baseline stereo cameras. It offers $165$ thousands annotated samples and additional $310$ thousands unlabeled samples and extends the effective perception range to [$-400$,$+400$] meters, focusing on highway long-range scenarios to stress perception capabilities beyond conventional benchmarks. $^*$NuPlan~\cite{nuplan} provides auto-labeled annotations.}
\vspace{-2mm}
\resizebox{\linewidth}{!}{
\begin{tabular}{l|llllcccc}
\toprule
\multirow{2}{*}{Dataset} &
\multirow{2}{*}{LiDARs} &
\multirow{2}{*}{Radars} &
\multirow{2}{*}{Cameras} &
\multirow{2}{*}{Localization} &
Sensor &
Synced &
Manually &
Effective \\
 &  &  &  &  & Count & Samples & Annotated & Range\\
 \hline
 KITTI \cite{geiger2012we} & 1x 64-beam & - & 2x RGB, 2x grayscale & GPS, IMU & \cellcolor{Snow4!16} 7 & \cellcolor{SkyBlue4!50} 216k & \cellcolor{DarkRed!5} 15k & \cellcolor{MediumPurple4!5} [0, +70] \\
 SemanticKITTI \cite{behley2019iccv} & 1x 64-beam & - & 2x RGB, 2x grayscale & GPS, IMU & \cellcolor{Snow4!16} 7 & \cellcolor{SkyBlue4!24} 43k & \cellcolor{DarkRed!15} 43k & \cellcolor{MediumPurple4!14} [-80, +80] \\
 ApolloScape \cite{wang2019apolloscape} & 2x & - & 2x RGB, 2x grayscale & GPS, IMU & \cellcolor{Snow4!18} 8 & \cellcolor{SkyBlue4!43} 143k & \cellcolor{DarkRed!7} 20k & \cellcolor{MediumPurple4!17} [-100, +100] \\
 A2D2 \cite{geyer2020a2d2audiautonomousdriving} & 5x 16-beam & - & 6x & GPS, IMU & \cellcolor{Snow4!29} 13 & \cellcolor{SkyBlue4!60} 392k & \cellcolor{DarkRed!14} 40k & \cellcolor{MediumPurple4!17} [-100, +100] \\
 H3D \cite{patil2019h3ddatasetfullsurround3d} & 1x 64-beam & - & 3x & GPS, IMU & \cellcolor{Snow4!14} 6 & \cellcolor{SkyBlue4!16} 27k & \cellcolor{DarkRed!9} 27k & \cellcolor{MediumPurple4!17} [-100, +100] \\
 Cityscapes 3D \cite{gählert2020cityscapes3ddatasetbenchmark} & - & - & 1x Stereo Pair & - & \cellcolor{Snow4!5} 2 & \cellcolor{SkyBlue4!15} 25k & \cellcolor{DarkRed!9} 25k & \cellcolor{MediumPurple4!17} [0, +200] \\
 Lyft L5 \cite{lyft2020} & 1x & - & 7x & - & \cellcolor{Snow4!18} 8 & \cellcolor{SkyBlue4!46} 170k & \cellcolor{DarkRed!10} 30k & \cellcolor{MediumPurple4!17} [-100, +100] \\
 A*3D \cite{astar-3d} & 1x 64-beam & - & 1x Stereo Pair & - & \cellcolor{Snow4!7} 3 & \cellcolor{SkyBlue4!22} 39k & \cellcolor{DarkRed!13} 39k & \cellcolor{MediumPurple4!17} [-100, +100] \\
 SeeingThoughFog \cite{bijelic2020seeingfogseeingfog} & 1x 64-beam, 1x 32-beam & 1x & 1x Stereo Pair, 1x Gated, 1x FIR & GPS,IMU & \cellcolor{Snow4!18} 8 & \cellcolor{SkyBlue4!5} 13.5k & \cellcolor{DarkRed!5} 13.5k & \cellcolor{MediumPurple4!21} [-120, +120] \\
 NuScenes \cite{caesar2020nuscenes} & 1x 32-beam & 5x & 6x & GPS, IMU & \cellcolor{Snow4!31} 14 & \cellcolor{SkyBlue4!60} 400k & \cellcolor{DarkRed!14} 40k & \cellcolor{MediumPurple4!17} [-100, +100] \\
 NuPlan \cite{nuplan} & 2x 20-beam, 3x 40-beam & - & 8x & GPS, IMU & \cellcolor{Snow4!33} 15 & \cellcolor{SkyBlue4!75} 62.5M & (4.3M$^*$) & \cellcolor{MediumPurple4!17} [-100, +100] \\
 NuImages \cite{caesar2020nuscenes} & - & - & 1x out of 6 & - & \cellcolor{Snow4!14} 6 & \cellcolor{SkyBlue4!36} 93k & \cellcolor{DarkRed!31} 93k & - \\
 Waymo - Perception \cite{sun2020scalability} & 1x mid-range, 4x short-range & - & 5x & - & \cellcolor{Snow4!22} 10 & \cellcolor{SkyBlue4!51} 230k & \cellcolor{DarkRed!65} 230k & \cellcolor{MediumPurple4!17} [-100, +100] \\
 Waymo - End2End \cite{xu2025wode2ewaymoopendataset} & - & - & 8x & - & \cellcolor{Snow4!18} 8 & \cellcolor{SkyBlue4!57} 321k & - & - \\
 ONCE \cite{mao2021one} & 1x 40-beam & - & 7x & - & \cellcolor{Snow4!18} 8 & \cellcolor{SkyBlue4!75} 1M & \cellcolor{DarkRed!6} 16k & \cellcolor{MediumPurple4!17} [-100, +100] \\
 AiMotive \cite{matuszka2022aimotivedataset} & 1x 64-beam & 2x & 2x RGB, 2x RGB Fisheye & GPS, IMU & \cellcolor{Snow4!16} 7 & \cellcolor{SkyBlue4!16} 26.5k & \cellcolor{DarkRed!9} 26.5k & \cellcolor{MediumPurple4!42} [-200, +200] \\
 Argoverse V2 \cite{Argoverse2} & 2x 32-beam & - & 1x Stereo Pair, 7x Ring Cameras & GPS & \cellcolor{Snow4!25} 11 & \cellcolor{SkyBlue4!44} 150k & \cellcolor{DarkRed!49} 150k & \cellcolor{MediumPurple4!46} [-250, +250] \\
 MAN TruckScenes \cite{fent2024mantruckscenesmultimodaldataset} & 6x & 6x & 4x & GPS, IMU & \cellcolor{Snow4!40} 18 & \cellcolor{SkyBlue4!18} 30k & \cellcolor{DarkRed!10} 30k & \cellcolor{MediumPurple4!42} [-226, +226] \\
\hline
 \textbf{TruckDrive (Ours)} & 7x long-range, 3x short-range & 10x & 1x / 3x Stereo Pair, 9x single & GPS, IMU & \cellcolor{Snow4!75} 37 & \cellcolor{SkyBlue4!59} 475k & \cellcolor{DarkRed!54} 165k & \cellcolor{MediumPurple4!75} [-400, + 400] \\
\bottomrule
\end{tabular}}
\label{tab:dataset_comparison}
\vspace{-2mm}
\end{table*}
%

%
\section{Related Work}
\label{sec:related_work}
Public vision datasets~~\cite{pascal-voc-2010,deng2009imagenet,lin2015microsoftcococommonobjects,zhou2017scene,zhou2019semantic,chang2015shapenetinformationrich3dmodel,wang-etal-2018-glue} have been a catalyst for progress in computer vision, providing a shared basis for developing and comparing novel algorithms. Autonomous driving has followed the same pattern where improvements in detection, prediction and planning have been tightly coupled to increasingly capable datasets.

\PAR{Early Autonomous Driving Datasets. } The field was pioneered by the KITTI dataset \cite{geiger2012we} and, later, its extensions \cite{Menze2018JPRS, Menze2015ISA, behley2019iccv, behley2020benchmarklidarbasedpanopticsegmentation, Liao2022PAMI}, among the firsts to provide synchronized camera and LiDAR data with $3D$ bounding boxes annotations. These datasets, however, are limited to a relatively small scale, ranges and scenarios.

\PAR{Large-Scale Multimodal Autonomous Driving Datasets. }
The next generation of datasets addressed these limitations by introducing $360$ degrees sensor coverage and a much larger scale. The nuScenes ecosystem \cite{caesar2020nuscenes} provides a full sensor suite for $3$D perception, which was later complemented by nuImages \cite{caesar2020nuscenes}, a large-scale dataset focused on $2$D object detection (OD), and nuPlan \cite{nuplan}, the first large-scale, real-world benchmark for motion planning. Similarly, the Waymo Open Dataset ecosystem \cite{sun2020scalability} offered an unprecedented scale. While initially focused on perception tasks, it has since expanded with the Waymo Motion dataset for trajectory forecasting \cite{Ettinger_2021_ICCV} and the Waymo E2E benchmark for evaluating end-to-end driving models. The Argoverse datasets \cite{Argoverse, Argoverse2} extended the common perception range up to $150$ meters and the Lyft Level $5$ dataset \cite{lyft2020} focused on providing large-scale HD maps. 
\vspace{-1mm}
\PAR{Task-Focused Autonomus Driving Datasets.} Along with the development of large-scale benchmarks, several datasets have made significant contributions by focusing on specific tasks and modalities. Cityscapes $3$D \cite{gählert2020cityscapes3ddatasetbenchmark} extended the popular semantic segmentation benchmark \cite{Cordts2015Cvprw, Cordts2016Cityscapes} with $3$D bounding box annotations, bridging the gap between $2$D and $3$D scene understanding. ApolloScape \cite{wang2019apolloscape} introduced a massive collection of data with a wide variety of tasks, including $3$D detection, lane segmentation, and dense trajectory information for simulation. Datasets from automotive OEMs, such as A2D2 (Audi) \cite{geyer2020a2d2audiautonomousdriving}, H3D (Honda) \cite{patil2019h3ddatasetfullsurround3d} and surround-view truck scenes from MAN Truckscenes \cite{fent2024mantruckscenesmultimodaldataset}, provide data from high-quality, industry-grade sensor configurations. KAIST dataset\cite{kim2020highwaydrivingdatasetsemantic} and aiMotive \cite{matuszka2022aimotivedataset} explored highway driving scenarios, although containing respectively only $1.2$k annotated frames and $12$k \textit{highway} frames. The ONCE dataset \cite{mao2021one} has pushed towards reducing annotation dependency by providing a large-scale benchmark for self-supervised learning, while A*3D \cite{astar-3d} and SeeingThroughFog \cite{bijelic2020seeingfogseeingfog} explored active learning strategies or novel sensor setups to improve annotation efficiency in highly challenging weather conditions. 

\PAR{Limits of Existing Autonomous Driving Datasets. }
As presented in Table \ref{tab:dataset_comparison}, prior autonomous driving datasets are dominated by urban, low speed setting and short effective ranges: $3$D labels are rarely present above $80$ meters, annotation density decrease rapidly with distance and long-range sensing is either absent or poorly represented \cite{Argoverse2, fent2024mantruckscenesmultimodaldataset}. Moreover, they often offer low annotation amount and sensor modalities are limited to a small set of cameras and short range LiDARs, pushing models to fit specific biases and leaving safe heavy-vehicle driving as an open challenge. 

\section{TruckDrive Dataset}
\label{sec:dataset}
We introduce \textit{TruckDrive}, a long-range, highway-focused dataset designed for heavy-vehicle autonomy. In this section, we first describe the TruckDrive domain and data collection process, emphasizing its diverse driving conditions, specialized sensor suite for high-speed perception and our cross-modal synchronization strategy. We further detail our annotation pipeline, which combines manual labeling with automated multi-view completion and kinematic refinement. Finally, we provide a quantitative analysis and comparison to foundational datasets (from Table~\ref{tab:dataset_comparison}), demonstrating gains in range, speed and trajectory coverage.

\subsection{Dataset Domain}
\label{sec:odd}
TruckDrive targets the driving domain of semi-trucks and other large commercial vehicles, covering scenarios that differ significantly from the urban, car-centric datasets commonly used in autonomous driving research. The dataset contains $3,828$ sequences recorded over $2$ years across $8$ U.S. states (NM, TX, VA, NC, TN, AR, WV, AZ), reaching a diversity-area metric \cite{sun2020scalability} of $1,261.3$ km$^2$ ($16.5\times$ WOD). Data collection spans all seasons ($48$\% fall, $32$\% winter, $15$\% spring, $5$\% summer) and diverse weather ($80$\% sunny/cloudy/overcast, $10$\% fog, $10$\% precipitation). Sequences last $15–25$ s with an average ego trajectory of $500$ m, comprising mainly highways ($3,244$), followed by extra-urban ($351$) and urban roads ($233$). Driving patterns include $45.8$\% cruise/accelerate/brake, $36.5$\% lane changes/overtakes, $5.4$\% close cut-ins, and $12.3$\% complex layouts (work zones, intersections, unprotected turns). Illumination coverage includes $3,285$ daytime, $367$ night, $122$ dusk, and $54$ dawn sequences.


\begin{figure}[t]
    \centering
       \includegraphics[width=\linewidth]{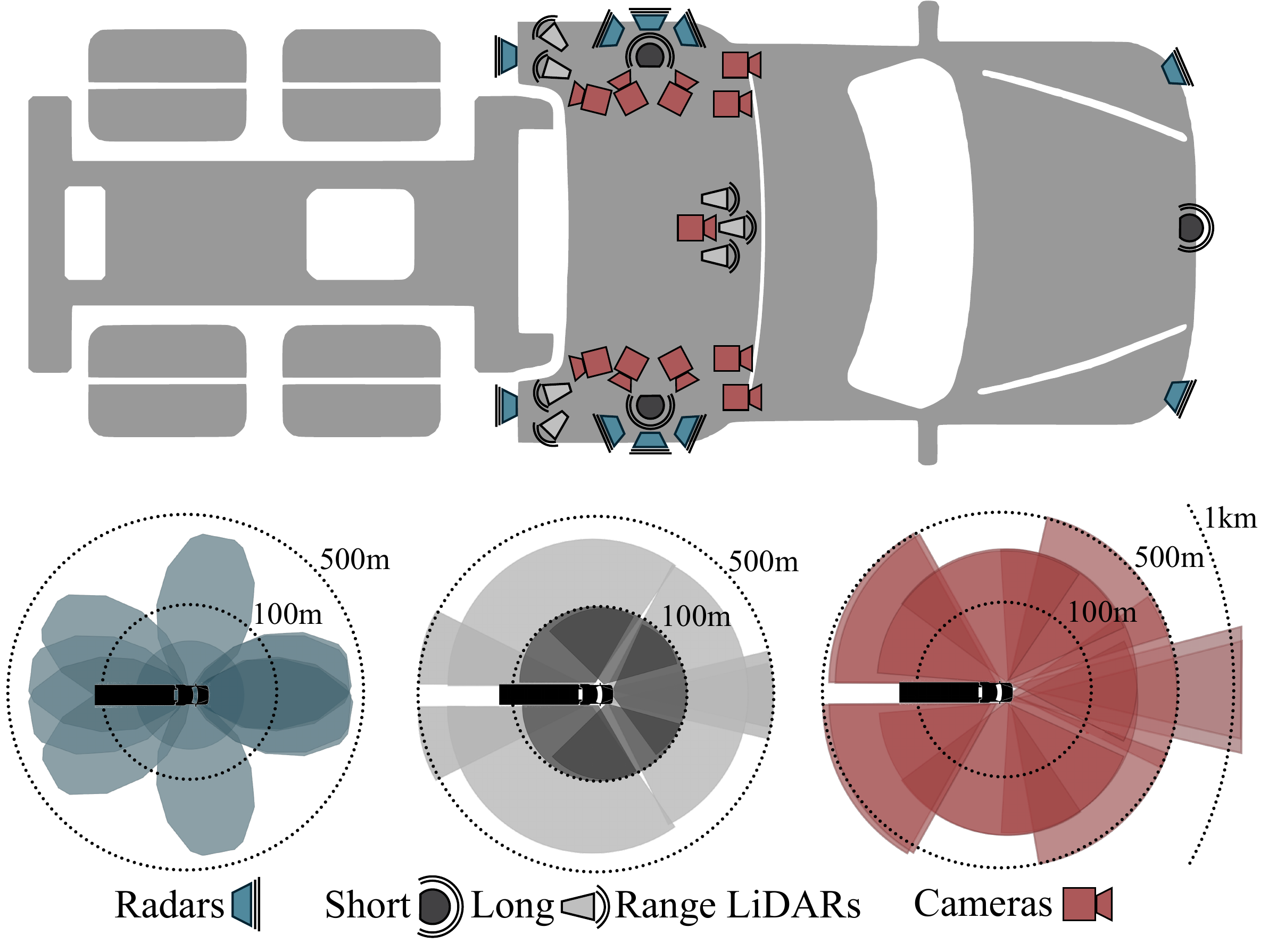}
       \vspace{-5mm}
    \caption{\textbf{Sensors Position and FoV.} Sensor position (top) and the nominal instrumented horizontal field of view (bottom) of, from left to right, radars, LiDARs and cameras, highlighting the unprecedented ranges at which they can operate.
    }
    \label{fig:sensors_fov}
    \vspace{-3mm}
\end{figure}
\newcolumntype{C}[1]{>{\centering\arraybackslash}m{#1}}

\begin{table}[t]
\centering
\renewcommand{\arraystretch}{1.3}
\caption{\textbf{Sensor Specifications and Raw Data Scale.} We present in detail our sensor platform, including RCCB cameras $3$D short-range (SR) LiDARs, a $4$D long-range (LR) FMCW LiDAR, and $4$D radars, capturing $475$ thousands synchronized frames}
\vspace{-2mm}
\label{tab:sensor_fused_rot}
\setlength{\tabcolsep}{-1pt} 

\begin{adjustbox}{max width=\linewidth}
\large
\begin{tabular}{l C{2.8cm} C{2.8cm} C{2.8cm} C{2.8cm}}
\toprule
& \textbf{Camera} & \multicolumn{2}{c}{\textbf{LiDAR}} & \textbf{Radar} \\
\cmidrule(lr){2-2}\cmidrule(lr){3-4}\cmidrule(lr){5-5}
&
\makecell{\textbf{RCCB}\\\textit{AR0820}} &
\makecell{\textbf{4D LR}\\\textit{Aeries II}} &
\makecell{\textbf{3D SR}\\\textit{OS0/OS1}} &
\makecell{\textbf{4D}\\\textit{ARS540}} \\
\midrule
\textbf{Make}       & OnSemi    & AEVA      & Ouster     & Continental \\
\textbf{Type}       & RCCB      & FMCW 4D   & 3D         & FMCW 4D \\
\textbf{Resolution} & $3848\times2168$ & $\sim$100 lines & $64/128\times 2048$ & --- \\
\textbf{FOV (H$\times$V)}        & $52.8^\circ\times28.9^\circ$ & $120^\circ\times30^\circ$ &
                      $360^\circ\times90^\circ/45^\circ$ & $\pm4^\circ$--$\pm20^\circ$ \\
\textbf{$f$ (Hz)}   & $5$--$10$ & \multicolumn{2}{c}{10} & 20 \\
\textbf{Raw Captures} & ~6.3M & \multicolumn{2}{c}{~7.8M} & ~6.0M \\
\textbf{Sync Timestamps} & 569k & \multicolumn{2}{c}{744k} & 601k \\
\midrule
\multicolumn{5}{c}{\textbf{Cross-Modal Sync Timestamps}: 475k} \\
\bottomrule
\end{tabular}
\end{adjustbox}

\vspace{-4mm}
\end{table}

\subsection{Long-Range Sensor Setup}
Our sensor suite, mounted on a semi-truck, is optimized for reliable perception in high-speed environments. Specifically, we employ $7$ FMCW LiDARs (AEVA Aeries II), capable of measuring up to $400$ meters and providing radial velocity, $3$ short-range LiDARs (Ouster OS0/OS1), to account for blind spots and objects very close to the ego and $10$ $4$D radars (Conti ARS540). Additionally, $11$ to $15$, depending on the configuration, RCCB cameras (9 short/medium focal and $1$ to $3$ long-focal stereo) provide high resolution imaging ($8MP$) at all ranges: QA verifies extrinsic accuracy below $0.015$\textdegree, bounding re-projection error beyond $200$m. We report placement and horizontal coverage in Figure \ref{fig:sensors_fov} and per-sensor specifications in Table \ref{tab:sensor_fused_rot}.

\PAR{FMCW Velocity.}
We rely on Frequency-Modulated Continuous-Wave (FMCW) technology, which allows to capture instantaneous radial velocity $v_r$ for each point in the point cloud. The velocity measurement is derived from the Doppler-induced phase shift  $\Delta \phi$ through
\vspace{-2mm}

\begin{equation}
v_r = \frac{\Delta \phi \cdot \lambda}{4\pi}\cos{\theta},
\label{eq:velocity}
\end{equation}

where $\lambda$ is the wavelength and $\theta$ the angle of incidence. 

\PAR{Geo-Inertial Poses (PPK).}
For accurate ego motion we fuse data from $2$ GNSS and $4$ IMUs in a tailored Post-Processing Kinematic (PPK) pipeline, yielding reliable global poses for synchronized frames. Rare failure cases are complemented with LiDAR SLAM~\cite{Pan_2024}, providing ground-truth trajectories suitable for precise localization.

\PAR{Sensors Synchronization}\label{sync} 
Each different sensor group is triggered and synced to a common clock, allowing no more than $5$ milliseconds between each unit capture. Cross-modal triggers are temporally aligned to enable near-simultaneous captures. Because our high-resolution cameras use a rolling shutter, showing a row-wise readout, aligning the other modalities to the image start time would induce a systematic temporal offset across rows. Instead, we define the reference timestamp at the image mid-exposure and synchronize LiDAR to this anchor 

\vspace{-2mm}
\begin{equation}
t_{\mathrm{ref}} = t_{\mathrm{img}}^{\mathrm{start}} + \tfrac{1}{2}T_{\mathrm{readout}},\quad \bigl|t_{\mathrm{LiDAR}} - t_{\mathrm{ref}}\bigr| \le 5\,\mathrm{ms},
\end{equation}

with a typical $T_{\mathrm{readout}}$ of $54$ milliseconds.

\subsection{Annotation}
We annotate $3$D cuboids through a three-stage pipeline that combines human annotation with automated label refinement. To maximize the richness of the annotated data, human annotators manually curate sequential frames containing complex interactions or edge cases; in total, more than $2000$ scenes are selected. Annotators then label $3$D cuboids and $2$D boxes and assign semantic classes. The selected annotations are subsequently refined automatically to enforce geometric and temporal consistency. For supervised learning tasks, the dataset provides around $140$\,k annotated training samples and $25$\,k annotated validation samples.

\begin{figure}[!t]
    \centering
    \begin{subfigure}[b]{0.48\linewidth}
        \centering
        \includegraphics[trim=0cm 0cm 0cm 0cm, clip, width=\linewidth]{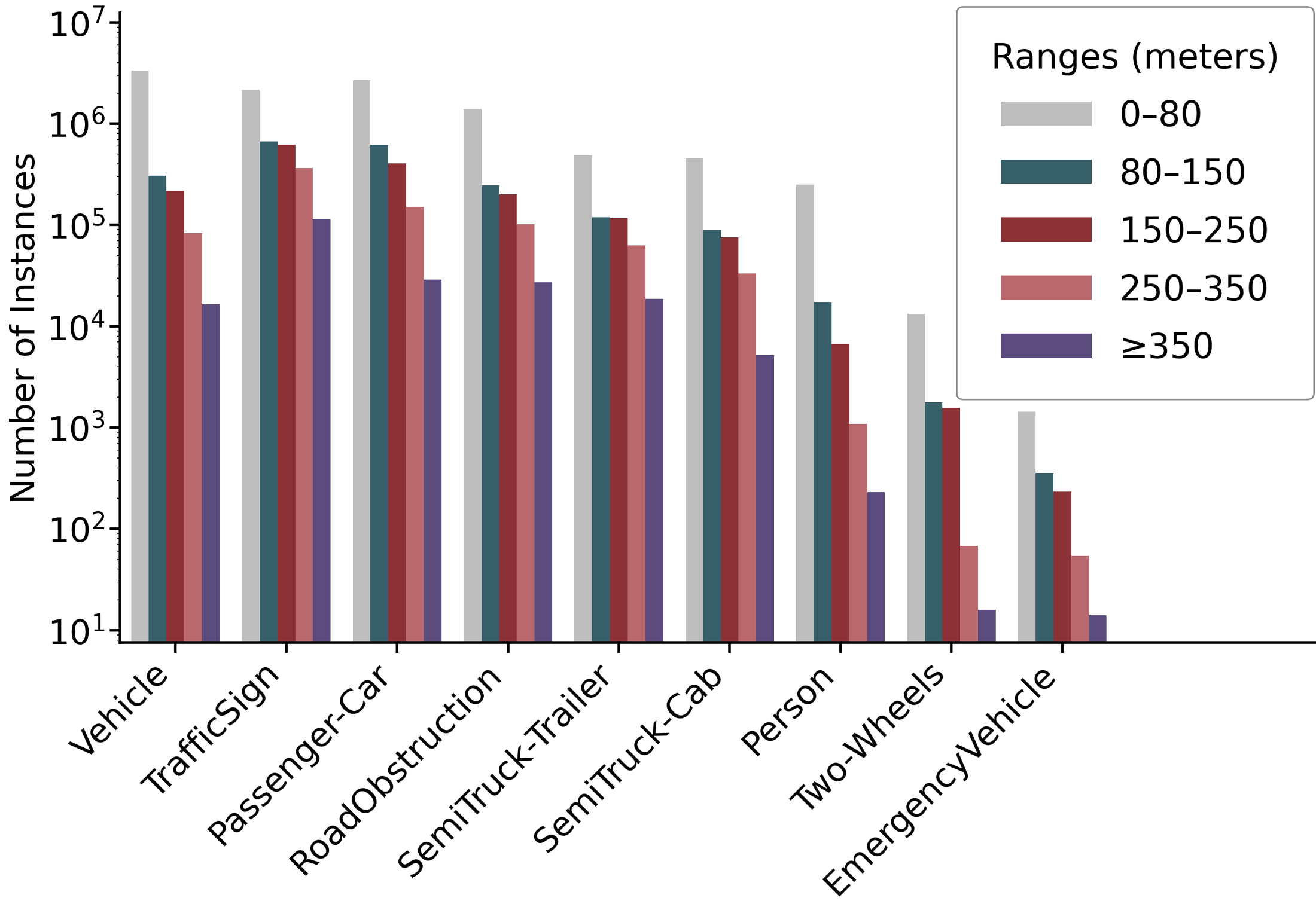}
        \caption{Class Labels Range Distribution}
        \label{fig:labels_dist}
    \end{subfigure}
    \hfill
    \begin{subfigure}[b]{0.48\linewidth}
        \centering
        \includegraphics[trim=0cm 0cm 0cm 0cm, clip, width=\linewidth]{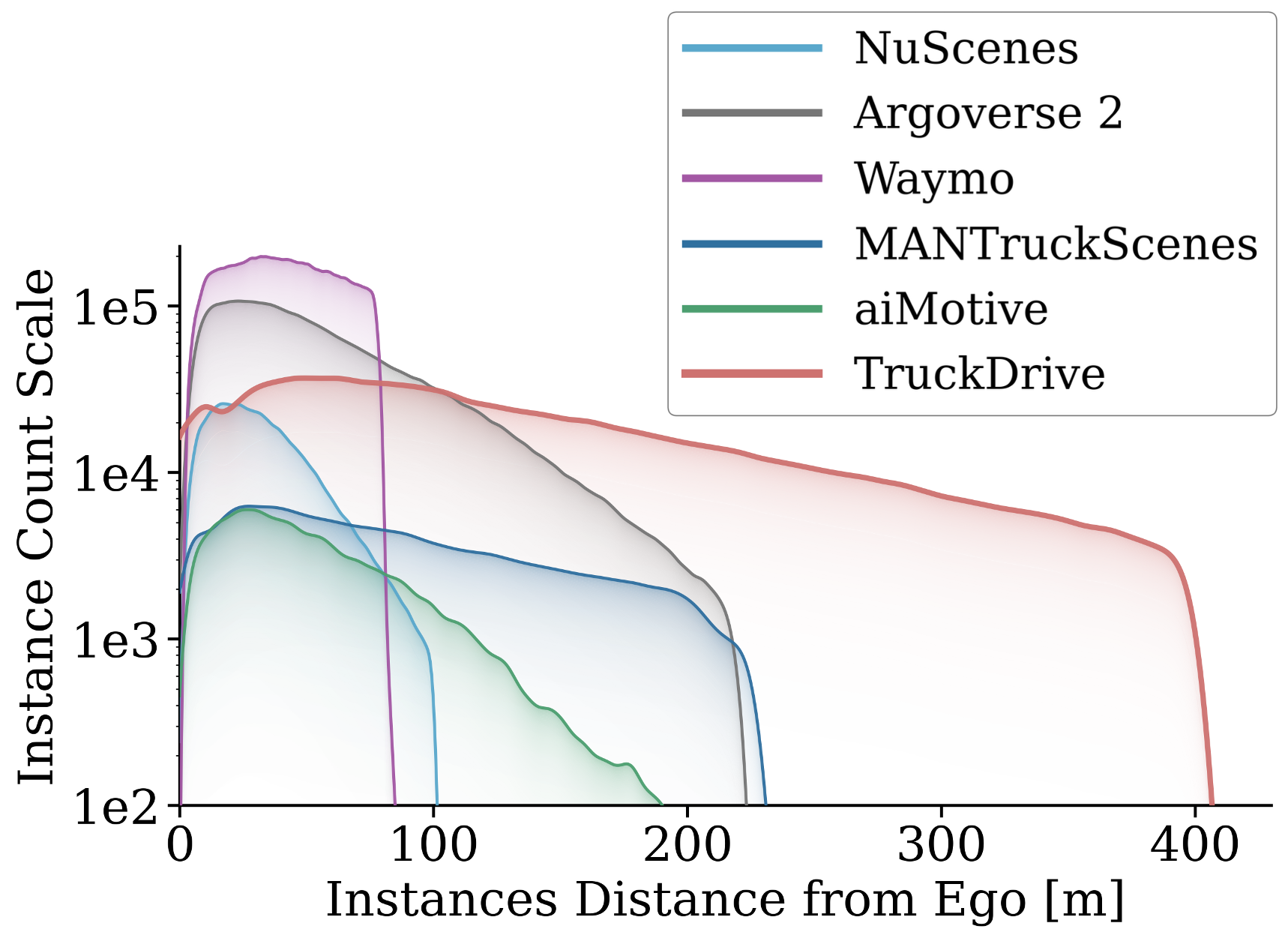}
        \caption{Instances Range Distribution}
        \label{fig:labels_dist_2}
    \end{subfigure}
    
    \vspace{2mm} 
    
    \begin{subfigure}[b]{0.48\linewidth}
        \centering
        \includegraphics[trim=0.2cm 0.2cm 0cm 0cm, clip, width=\linewidth]{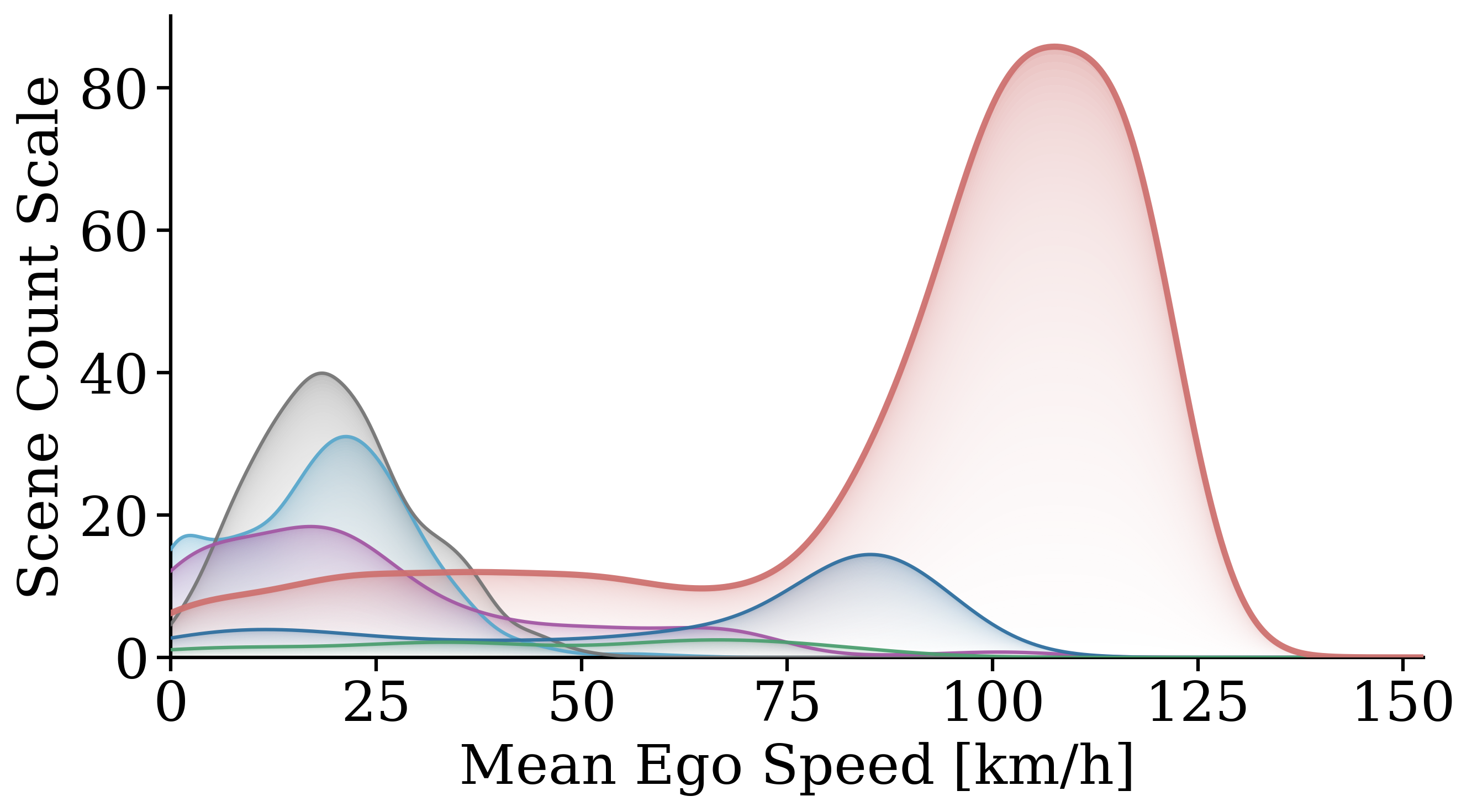}
        \caption{Ego Speed Distribution}
        \label{fig:speed_dist}
    \end{subfigure}
    \hfill
    \begin{subfigure}[b]{0.48\linewidth}
        \centering
        \includegraphics[trim=0.2cm 0.2cm 0cm 0cm, clip, width=\linewidth]{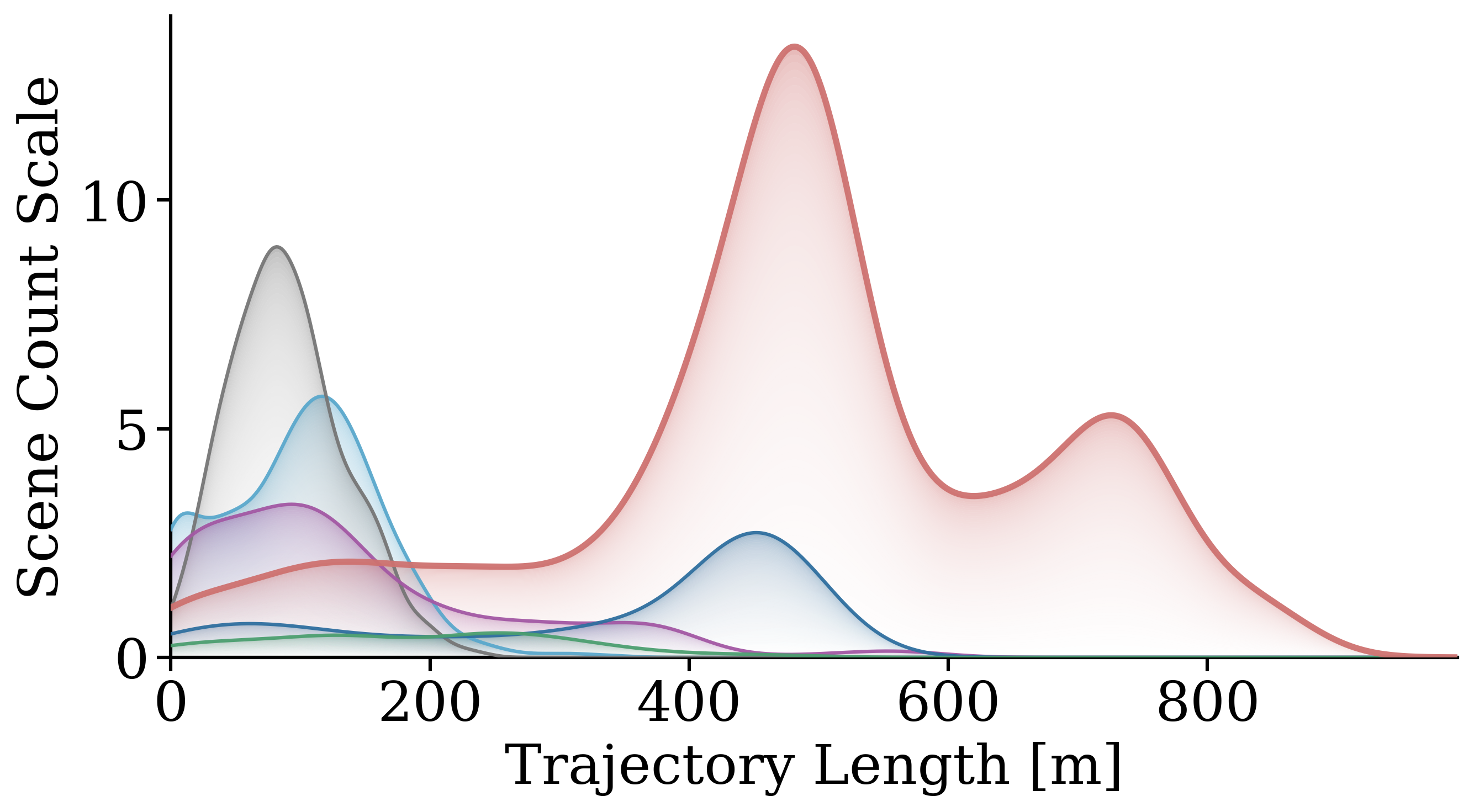}
        \caption{Scene Length Distribution}
        \label{fig:scene_length}
    \end{subfigure}
    
    \vspace{-3mm}
    \caption{\textbf{Dataset Analysis. }Our dataset comprises an unprecedented density of instance objects at ranges (greater than $200$ meters) yet to be explored in publicly available datasets (a,b), as well as driving speeds $5$ times higher (c) and sequences with traveled length up to $8$ times longer (d) than existing benchmarks.}
    \label{fig:combined_2x2}
    \vspace{-5mm}
\end{figure}

\PAR{Stage 1: Human Annotation Primitives.} During this stage, annotators produce geometric primitives consisting of $3$D cuboids and $2$D boxes (with relative Occlusion and Truncation parameters) and assign semantic labels to all identified objects. $3$D boxes are then iteratively adjusted using their projection into the cameras to reduce offset and avoid “ghost” objects. The annotation procedure results in $85$ classes which we regroup in $9$ main categories as shown in Figure \ref{fig:labels_dist}. The $9$ classes to be captured are \textit{traffic signs}, \textit{passenger cars}, all types of road debris and interferences such as lost cargo, potholes, and cones (collectively referred to as \textit{road obstructions}), \textit{humans}, \textit{semi-trucks} in both their cabins and trailers, \textit{$2$-wheeled vehicles}, \textit{emergency vehicles} like police cars, ambulances or road-construction vehicles that can halter the nominal planning behavior and \textit{vehicles} of different sizes, from heavy-duty vehicles, buses or single unit trucks to RV, trailers and equipment. Vulnerable Road Users are identified and included in the coarser categories.

\PAR{Stage 2: Primitive Augmentation. }For each timestamp we project the initial $3D$ cuboids into all camera views and match them against detections from a $2D$ object detector, by solving a bipartite assignment (Hungarian algorithm) with Intersection-over-Union as cost matrix. When a $2D$ detection has no correspondence, we fall back to the geometric projection or an existing $2D$ label. We handle truncations and perform class-wise Non Maximum Suppression (NMS) to promote high confidence $2D$ detections, resulting in the set of matched 3D detections and 2D-only candidates.

\PAR{Stage 3: Refinement and Completion. }The existing matched $3D$ annotations are transformed into a global coordinate frame and their trajectories are refined through a kinematically constrained optimization, enforcing plausible motion and reducing yaw jitter. Specifically we minimize
\vspace{-2mm}

\begin{fleqn}[0pt]
\begin{align}
\label{eq:traj_obj}
\min_{\{s_t^k,d_t^k\}} \sum_{t\in\mathcal{T}_k}
\big[\lambda_{\text{o}}L^{\text{o}}_t
+\lambda_{\psi}L^{\psi}_t
+\lambda_{d}L^{d}_t
+\lambda_{\text{smooth}}L^{\text{sm}}_t\big],\\
\label{eq:loss_obspsi}
L^{\text{o}}_t=\rho(\|c(s_t^k)-\hat c_t^k\|_2),\quad
L^{\psi}_t=\rho(\mathrm{ang}(\psi_t^k,\hat\psi_t^k)),\\
\label{eq:loss_dsm}
L^{d}_t=\rho(\|d_t^k-\hat d_t^k\|_2),\quad
L^{\text{sm}}_t=\|\Delta v_t^k\|_2^2+\|\Delta^2\psi_t^k\|_2^2.
\end{align}
\end{fleqn}
subject to a unicycle model
\vspace{-2mm}
\begin{equation}
\label{eq:unicycle}
\begin{aligned}
x_{t+1}^k &= x_t^k + \Delta t\, v_t^k \cos\psi_t^k, &
\psi_{t+1}^k &= \psi_t^k + \Delta t\, \omega_t^k, \\
y_{t+1}^k &= y_t^k + \Delta t\, v_t^k \sin\psi_t^k,&
\kappa_t^k &= \omega_t^k \slash  v_t^k.
\end{aligned}
\end{equation}

Here, \(s_t^k=(x_t^k,y_t^k,\psi_t^k,v_t^k,\omega_t^k)\) is the per-track state, \(d_t^k=(\ell_t^k,w_t^k,h_t^k)\) are box sizes, \(c(s_t^k)=(x_t^k,y_t^k)\) extracts the box center, hats \(\hat{\cdot}\) denote noisy estimates, \(\rho(\cdot)\) is a robust loss (Huber with scale \(\delta_\rho\)), \(\operatorname{ang}\) is the angle difference, $\Delta$/$\Delta^2$ are first and second finite differences.

For short gaps $t\in[t_1,t_2]$ with missing frames we initialize bounding boxes by interpolating
\vspace{-2mm}

\begin{equation}
\label{eq:gap_fill}
\begin{aligned}
\tilde c_t&=(1-\alpha)c_{t_1}+\alpha c_{t_2},\quad
\tilde \psi_t=\operatorname{slerp}(\psi_{t_1},\psi_{t_2};\alpha),\\
\tilde d_t&=(1-\alpha)d_{t_1}+\alpha d_{t_2},\quad
\alpha=(t-t_1) \slash (t_2-t_1),
\end{aligned}
\end{equation}

then refine jointly using Equations \eqref{eq:traj_obj} to \eqref{eq:unicycle}. 

Concurrently, we lift unmatched $2D$ candidates from Stage $2$ into $3D$. For each camera $c$, we project the eight cuboid corners of a 3D hypothesis $p=(x,y,z,\ell,w,h,\psi)$ and form the tight axis-aligned 2D box $\hat b_c(p)$. 
We retain only those camera views whose Stage $2$ detection $b_{c,t}=[x_0,y_0,x_1,y_1]$ has sufficient overlap with the hypothesis, defined as $\mathrm{IoU}\!\big(\hat b_c(p),\,b_{c,t}\big)\ge 0.3$, and optimize $p$ so that the projected boxes fit the detections across the retained views                                     
\begin{equation}
\label{eq:mv_obj}
\sum_{c\in\mathcal{C}}
\Big[\lambda_{\mathrm{iou}}\big(1-\mathrm{IoU}(\hat b_c(p),b_c)\big)
+\lambda_g\big(z_{\min}(p)-z_g\big)^2\Big],
\end{equation}

where $z_g$ is the local ground height from the accumulated LiDAR map. $3$D objects are then tracked over time with a offline tracker~\cite{wang2021immortal}, identity-aligned to ground truth via temporal IoU voting and merged with the smoothed ground-truth boxes to form the final annotation set.

\subsection{Dataset Analysis}
TruckDrive, compared in Table \ref{tab:dataset_comparison} with other benchmarks, introduces an unprecedented sensing configuration with $37$ heterogeneous sensors, double the number available in the second most sensor-rich dataset ($18$), enabling full $360^\circ$ perception coverage with both long and short-range redundancy and enhancing robustness in complex environments. TruckDrive’s LiDAR extends up to $400$ meters in both the forward and rear directions, twice the maximum range reported in previous benchmarks ($220$\,m). The dataset comprises approximately $165,000$ manually annotated frames, which is comparable in scale to the largest publicly available datasets ($230$\,k). Per-class instances are distributed uniformly across the full perception range, yielding balanced near and far-field samples (Fig.~\ref{fig:labels_dist}). The density of annotated $3$D boxes decays gradually with distance up to $400$\,m (Fig.~\ref{fig:labels_dist_2}), while $2$D boxes extend well beyond $1000$\,m, in contrast to prior urban-focused datasets~\cite{caesar2020nuscenes, sun2020scalability, Argoverse2} where annotations beyond $100-200$\,m are rare and instance density drops sharply after $80$\,m.
Figures~\ref{fig:speed_dist} and \ref{fig:scene_length} highlight highway dynamics in TruckDrive. Speeds span from low on/off ramp segments to up to $130$\,km/h, surpassing urban datasets capped below $75$\,km/h. Sequences extend to $900$\,m (against $400$\,m of urban datasets), enabling temporal reasoning at high speed and more faithful evaluation of long-horizon perception and prediction.

\section{Driving Tasks and Challenges} 

\begin{figure*}[t]
    \centering
    \vspace{-2mm}
    \includegraphics[trim=0.4cm 0cm 0cm 0cm, clip, width=\textwidth]{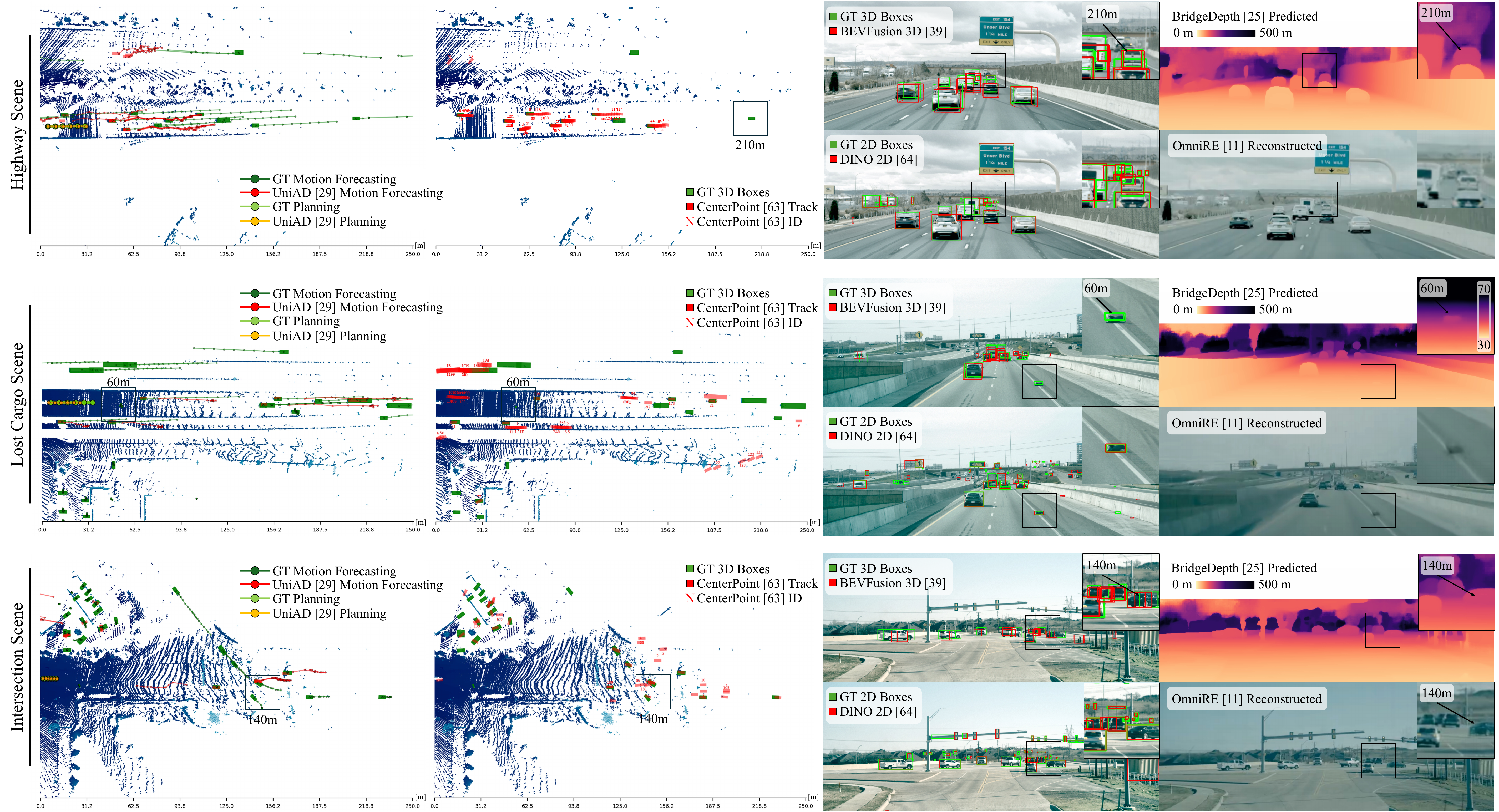}
    \vspace{-5mm}
    \caption{\textbf{Driving Tasks and Challenges. }We report qualitative results of the best baselines across planning, 2D/3D object detection, depth estimation and scene reconstruction. Even when trained on TruckDrive, existing methods struggle in the long-range, high-speed regime. Planning modules exhibit conservative behavior due to low-speed assumptions. Grid-based BEV models degrade perception as large spatial coverage demands heavy downsampling, erasing safety-critical details such as small debris or lost cargo \cite{liu2022bevfusion}, while depth methods struggle beyond 200m and in sky regions \cite{guan2025bridgedepth}, revealing limited distance awareness and motivating architectures for highway-scale perception.}
    \label{fig:qualitative_resuls}
    \vspace{-3mm}
\end{figure*}

\label{sec:tasks}
We use the proposed dataset at hand to evaluate recent perception and driving methods across typical tasks, such as $2$D and $3$D object detection, tracking, depth estimation, LiDAR forecasting, moving object segmentation, $3$D scene reconstruction and end-to-end planning.
This evaluation investigates whether current state-of-the-art approaches, primarily developed and optimized for urban driving datasets, can generalize to the speed, long-range and large-scale highway scenarios present in TruckDrive. To this end, all tested models have been trained on our TruckDrive data.
We train all models with a consistent train-validation split made of $140$ and $25$ thousand samples respectively and follow standard metrics and protocols. We couple quantitative with qualitative results for the target domain in Figure \ref{fig:qualitative_resuls}.

\subsection{2D Object Detection} 
In nuScenes and KITTI \cite{caesar2020nuscenes,geiger2012we}, $2$D performance is largely driven by $3$D detectors due to low image resolution and wide FOV; $3$D NMS in lifted space handles occlusion better than image-space NMS. At kilometer ranges, however, objects in those benchmarks would be sub-pixel, whereas our $8$MP imagery keeps them resolvable, so only $2$D detectors are able detect them. We train state-of-the-art architectures \cite{yolo11, detr, vitdet, 2022dino} and report results in Table \ref{tab:2d_detection}.

\subsection{3D Object Detection}
We evaluate long-range $3$D object detection using three SOTA models on our dataset, spanning a LiDAR based model \cite{bai2022transfusionrobustlidarcamerafusion}, a camera-based method \cite{jiang2023far3dexpandinghorizonsurroundview} and a common LiDAR-camera fusion architecture \cite{liu2022bevfusion}. We report average precision over three range bins in Table \ref{tab:3d_object_detection}.

\subsection{3D Multi Object Tracking}
We evaluate whether state-of-the-art tracking methods can handle the long-horizon scenes and high differential velocities between the ego and other agents in TruckDrive, which stress association over long gaps and occlusions. We report MOT results for a query based approach~\cite{zhang2022mutr3d} and two $3$D boxes based methods ~\cite{wang2021immortal, yin2021centerbased3dobjectdetection} in Table~\ref{tab:mot_methods}.

\begin{table}[t]
\centering
\caption{\textbf{2D Object Detection Results.}
We follow CoCo \cite{lin2015microsoftcococommonobjects} and report mean average precision (mAP) at $0.50$ IoU, mAP at $0.75$ IoU, and mAP at short ($0-50$\,m, SR), medium ($50-150$\,m, MR), long ($150-250$\,m, LR), and ultra-long-range ($250$+, UR).}
\vspace{-2mm}
\label{tab:2d_detection}
\resizebox{\linewidth}{!}{
\begin{tabular}{lccccccc}
\toprule
\textbf{Method} & \textbf{mAP}$\uparrow$ & 
\textbf{mAP$_{50}$}$\uparrow$ & 
\textbf{mAP$_{75}$}$\uparrow$ & 
\textbf{mAP$_{SR}$}$\uparrow$ &
\textbf{mAP$_{MR}$}$\uparrow$ & 
\textbf{mAP$_{LR}$}$\uparrow$ & 
\textbf{mAP$_{UR}$}$\uparrow$ \\ 
\midrule
DETR   \cite{detr}  & 12.70\% & 23.90\% & 12.20\% & 41.20\% & 24.70\% & 8.90\% & 1.00\% \\
ViTDet \cite{vitdet}  &  27.30\% & 37.60\% & 30.80\% & 58.30\% & 51.80\% & 33.90\% & 3.30\% \\
YOLO11x    \cite{yolo11}   & 28.90\%  & 39.00\%  & 31.60\% & 36.30\% & 29.40\% & 8.20\%  & 2.00\% \\
DINO      \cite{2022dino}   & 37.80\%  &  54.20\% & 40.30\% & 63.90\% & 54.60\% & 43.20\% &  15.30\% \\
\bottomrule
\end{tabular}}
\vspace{-2mm}
\end{table}

\begin{table}[t]
    \centering
    \caption{\textbf{3D Object Detection Results}. We report mAP for $3$ baselines using a single or a combination of LiDAR (L) and Camera (C) data, divided into short ($0-50$\,m, SR), medium ($50-150$\,m, MR), long ($150-250$\,m, LR) and full detection ranges.}
    \vspace{-2mm}
    \resizebox{\linewidth}{!}{
    \addtolength{\tabcolsep}{+0.2em}
    \begin{tabular}{lc|c|ccc}
    \toprule
        \multirow{ 2}{*}{\textbf{Method}} & \multirow{ 2}{*}{\textbf{Mode}}& \textbf{mAP}$\uparrow$ & \textbf{mAP}$\uparrow$ & \textbf{mAP}$\uparrow$ & \textbf{mAP}$\uparrow$ \\
        && Full & SR & MR & LR \\
        \midrule
        {Far3D \cite{jiang2023far3dexpandinghorizonsurroundview}} & C &14.04\%&35.54\%&11.07\%&0.33\%  \\
        {TransFusion-L} \cite{bai2022transfusionrobustlidarcamerafusion} & L &25.24\%&30.12\%&22.25\%&22.25\%\\
        {BEVFusion} \cite{liu2022bevfusion} & L+C &  26.45\%&32.32\%&22.77\%&22.69\%\\ 
    \bottomrule
    	\end{tabular}
    \label{tab:3d_object_detection}}
\end{table}

\begin{table}[t]
\centering
\caption{\textbf{3D Multi Object Tracking Results}. We report AMOTA, AMOTP and Recall for a query based and two LiDAR based methods.{$^\dagger$} uses \cite{liu2022bevfusion} inference detections.}
\vspace{-2mm}
\resizebox{\linewidth}{!}{
        \begin{tabular}{l c| ccc}
        \toprule
            \textbf{Method} & \textbf{Mode} & \textbf{AMOTA}$\uparrow$ & \textbf{AMOTP}$\downarrow$ & \textbf{Recall}$\uparrow$ \\
            \midrule
            {MUTR3D \cite{zhang2022mutr3d}} & Query & 6.1\% & 79.0\% & 11.4\% \\
            {Immortal Tracker {$^\dagger$} \cite{wang2021immortal}} & 3D Box & 12.8\% & 77.2\% & 20.7\% \\
            {CenterPoint {$^\dagger$} \cite{yin2021centerbased3dobjectdetection}} & 3D Box & 13.0\% & 76.9\% & 21.5\%\\
        \bottomrule
        \end{tabular}
        }
\label{tab:mot_methods}
\vspace{-3mm}
\end{table}

\subsection{Depth Estimation}
We train monocular, stereo and surround depth estimation models under long-range LiDAR supervision to assess the capability of current approaches in the TruckDrive domain. 
For all subtasks, we report standard task metrics alongside unified, distance-binned depth metrics, ensuring balanced evaluation across ranges and avoiding the near-range bias and limited  range-dependent interpretability of disparity-based or relative-error metrics.

\PAR{Depth Evaluation Ground-Truth.} For our benchmark, we build dense LiDAR ground truth by accumulating static points and filtering dynamic objects through the FMCW capabilities of our $4$D LiDAR. The resulting depth map is projected into each frame, where we reintroduce dynamic points based on their timestamps, filter out view-dependent occlusions and enhance temporal consistency using dense depth priors inferred from an ensemble of depth foundation models. Additional details in the Supplementary Material.

\PAR{Surround Views.} Leveraging the wide, calibrated overlap among five high-resolution cameras arranged to ensure extensive, overlapping surround coverage, we train two state-of-the-art models \cite{keetha2025mapanything, schmied2023r3d3} for metric surround depth estimation and report results in Table~\ref{tab:ddad_sota}, evaluated against the dense LiDAR ground truth. Task-specific relative metrics are reported following \cite{schmied2023r3d3}.

\PAR{Stereo Views.} The forward-facing cameras are arranged in a wide-baseline stereo configuration (approx. $1.57$\,m), providing a strong geometric basis for depth perception via triangulation. We evaluate state-of-the-art learning-based stereo matching methods \cite{guan2024neural, guan2025bridgedepth, cheng2025monsterunifiedstereomatching} and report results in Table~\ref{tab:stereo_depth}. We report task-specific disparity metrics following the KITTI stereo benchmark \cite{Menze2015ISA, Menze2018JPRS}.

\PAR{Monocular View.} We benchmark recent existing monocular depth estimation models \cite{piccinelli2025unidepthv2, bhat2023zoedepthzeroshottransfercombining, hu2024metric3dv2}, which infer depth from single images without geometric priors, to assess their ability to generalize to the scale and appearance of distant objects. Each model is trained twice: once using the same $5$ cameras employed for surround views, and once using the left stereo camera, enabling direct comparison with stereo and surround-view architectures. Results are reported in Table~\ref{tab:depth_estimation}. Task-specific metrics are reported following the KITTI benchmark \cite{Uhrig2017THREEDV} for monocular depth estimation. 

\begin{table}[t]
\centering
\caption{\textbf{Depth Estimation Results}. We report performances for surround (a), stereo (b) and monocular (c) depth estimation tasks. 
Each method is evaluated with standard accuracy and error metrics at short (0-50m, SR), medium (50-150m, MR), long (150-250m, LR) and ultra (250-1000m, UR) range bins.}
\vspace{-2mm}

\begin{subtable}[t]{0.45\textwidth}
    \centering
    \caption{Multi-Camera Surround Depth Estimation}
    \resizebox{\linewidth}{!}{
    \begin{tabular}{lcccc|cccc}
    \toprule
        & \multicolumn{4}{c}{\textbf{Distance-Binned MAE (Depth)}} & \multicolumn{4}{c}{\textbf{Task-Specific Depth Metrics}} \\
        \cmidrule(lr){2-5} \cmidrule(lr){6-9}
        \textbf{Method}
        & \textbf{SR} $\downarrow$
        & \textbf{MR} $\downarrow$
        & \textbf{LR} $\downarrow$
        & \textbf{UR} $\downarrow$
        & \textbf{Abs Rel} $\downarrow$
        & \textbf{Sq Rel} $\downarrow$
        & \textbf{RMSE} $\downarrow$
        & $\boldsymbol{\delta_{1}}$ $\uparrow$ \\
        \midrule
        R3D3 \cite{schmied2023r3d3} 
        & 7.99 & 25.35 & 74.02 & 181.22 & 0.30 & 0.21 & 37.60 & 0.49 \\
        MapAnything \cite{keetha2025mapanything} 
        & 5.05 & 16.73 & 39.19 & 121.15  & 0.19 & 6.13 & 26.40 & 0.73 \\
    \bottomrule
    \end{tabular}}
    \label{tab:ddad_sota}
    \vspace{-2mm}
\end{subtable}%

\hspace{0.05\textwidth} 
\begin{subtable}[t]{0.45\textwidth}
    \centering
    \caption{Stereo Disparity Estimation}
    \label{tab:stereo_depth}
    \setlength{\tabcolsep}{6pt}
    \resizebox{\linewidth}{!}{
    \begin{tabular}{lcccc|ccc}
    \toprule
        & \multicolumn{4}{c}{\textbf{Distance-Binned MAE (Depth)}} & \multicolumn{3}{c}{\textbf{Task-Specific Disparity Metrics}} \\
        \cmidrule(lr){2-5} \cmidrule(lr){6-8}
        \textbf{Method} 
        & \textbf{SR} $\downarrow$
        & \textbf{MR} $\downarrow$
        & \textbf{LR} $\downarrow$
        & \textbf{UR} $\downarrow$
        & \textbf{D1-bg} $\downarrow$
        & \textbf{D1-fg} $\downarrow$
        & \textbf{D1-all} $\downarrow$ \\
    \midrule
        NMRF \cite{guan2024neural}        & 3.39 & 9.13 & 20.92 & 40.88  & 26.98 & 21.95 & 21.95 \\
        MonSter++ \cite{cheng2025monsterunifiedstereomatching} & 4.41 & 9.21 & 21.39 & 62.18 & 26.09 & 23.94 & 29.07 \\
        BridgeDepth \cite{guan2025bridgedepth} & 2.53 & 8.34 & 20.21 & 69.10 & 28.74 & 11.12 & 28.57 \\
    \bottomrule
    \end{tabular}}
    \vspace{-2mm}
\end{subtable}

\hspace{0.05\textwidth} 
\begin{subtable}[t]{0.47\textwidth}
    \centering
    \addtolength{\tabcolsep}{-0.2em}
    \caption{Monocular Depth Estimation}
    \resizebox{\linewidth}{!}{
    \begin{tabular}{lcccc|cccc}
    \toprule
        & \multicolumn{4}{c}{\textbf{Distance-Binned MAE (depth)}} 
        & \multicolumn{4}{c}{\textbf{Task-Specific Depth Metrics}} \\
        \cmidrule(lr){2-5} \cmidrule(lr){6-9}
        \textbf{Method} 
        & \textbf{SR} $\downarrow$
        & \textbf{MR} $\downarrow$
        & \textbf{LR} $\downarrow$
        & \textbf{UR} $\downarrow$
        & \textbf{SILog}$\downarrow$ 
        & \textbf{sqERel}$\downarrow$ 
        & \textbf{absERel}$\downarrow$ 
        & \textbf{iRMSE}$\downarrow$ \\
    \midrule
            \multicolumn{9}{l}{\textbf{Multi-View}} \\
    \midrule
        ZoeDepth \cite{bhat2023zoedepthzeroshottransfercombining} & 4.77 & 17.63 & 44.00 & 114.30 & 27.25 & 0.13 & 0.16 & 8.54 \\
        Metric3Dv2 \cite{hu2024metric3dv2}   & 4.68 & 15.26 & 42.11 & 144.51 & 25.31 & 0.11 & 0.17 & 9.06 \\                     
        UniDepthv2 \cite{piccinelli2025unidepthv2} & 3.52 & 12.30 & 28.63 & 103.94 & 21.07 & 0.06 & 0.14 & 2.85 \\
    \midrule
        \multicolumn{9}{l}{\textbf{Single-View}} \\
    \midrule
        ZoeDepth \cite{bhat2023zoedepthzeroshottransfercombining} & 4.15 & 15.80 & 45.55 & 133.78 & 23.93 & 0.07 & 0.20 & 3.51 \\
        Metric3Dv2 \cite{hu2024metric3dv2}                       & 3.28 & 12.81 & 27.53 & 94.47 & 22.01 & 0.05 & 0.14 & 2.75 \\
        UniDepthv2 \cite{piccinelli2025unidepthv2} & 2.66 & 10.63 & 28.37 & 102.58 & 20.08 & 0.05 & 0.13 & 2.45 \\
    \bottomrule
    \end{tabular}}
    \label{tab:depth_estimation}
\end{subtable}
\vspace{-7mm}
\end{table}

\subsection{Temporal Scene Modeling and Reconstruction}
Predicting future scene geometry is fundamental for safe motion planning. We benchmark recent methods on the LiDAR forecasting task over a challenging $250$ meters Region Of Interest (ROI) ahead of the ego vehicle, comparing a LiDAR-only \cite{khurana2023point}, a camera-only \cite{yang2023vidar}, and a multi-modal fusion network \cite{PalladinAndBruckerLRS4Fusion}. We report range-binned results in Table~\ref{tab:lidar_forecasting}.
For dynamic modeling, we evaluate a LiDAR-based moving-object segmentation method \cite{mersch2022ral} chosen for its strong out-of-domain generalization. As shown in Table~\ref{tab:evaluation_mos}, the pretrained model struggles at longer distances, indicating the need for long-range training to improve detection.
Beyond discrete object-level tasks, high-fidelity scene reconstruction on long-range data is critical for photorealistic digital twins and dense scene understanding. Therefore, we assess a Neural Radiance Fields (NeRF) \cite{nerfstudio} and two $3$D Gaussian Splatting (3DGS) methods \cite{chen2025omnire, Zhou_2024_CVPR} in Table~\ref{tab:nerfgs}. 

\begin{table}[t]
\centering
\caption{\textbf{LiDAR Forecasting Results}. We evaluate single and multi-modal state of the arts methods with Chamfer Distances (CD) of $1$ and $3$ seconds and L1 error.}
\vspace{-2mm}
\label{tab:lidar_forecasting}

\setlength{\tabcolsep}{8pt}
\resizebox{\linewidth}{!}{
\begin{tabular}{l l c cc cc}
\toprule
\multirow{2}{*}{\makecell{\textbf{History} \\ \textbf{Horizon}}} & \multirow{2}{*}{\textbf{Method}} & \multirow{2}{*}{\textbf{Modality}} & \multicolumn{2}{c}{\textbf{1s}} & \multicolumn{2}{c}{\textbf{3s}} \\
 & & & CD $\downarrow$ & L1 (m) $\downarrow$ & CD $\downarrow$ & L1 (m) $\downarrow$ \\
\midrule
\multirow{3}{*}{1s} 
 & 4DOcc \cite{khurana2023point} & L & 18.93 & 4.69 & - & - \\
 & ViDAR \cite{yang2023vidar} & C & 58.23 & 20.21 & 51.72 & 20.69 \\
 & LRS4Fusion \cite{PalladinAndBruckerLRS4Fusion} & L + C & 15.82 & 3.31 & 39.03 & 3.82 \\
\midrule
\multirow{3}{*}{3s} 
 & 4DOcc \cite{khurana2023point} & L & 23.58 & 3.00 & 47.81 & 4.29 \\
 & ViDAR \cite{yang2023vidar} & C & 57.28 & 20.14 & 56.20 & 20.53 \\
 & LRS4Fusion \cite{PalladinAndBruckerLRS4Fusion} & L + C & 16.38 & 2.49 & 42.93 & 4.05 \\
\bottomrule
\end{tabular}}
    \vspace{-3mm}
\end{table}

\begin{table}[!t]
\centering
\addtolength{\tabcolsep}{-0.25em}
\caption{\textbf{3D Moving Object Segmentation Results}. $^\ddagger$ indicates results from the public KITTI \cite{geiger2012we} checkpoint.}
\vspace{-2mm}
\scriptsize
\begin{tabular}{l*{8}{c}}
\toprule
\multirow{2}{*}{\textbf{Method}} & \multicolumn{2}{c}{\textbf{SR}} & \multicolumn{2}{c}{\textbf{MR}} & \multicolumn{2}{c}{\textbf{LR}} & \multicolumn{2}{c}{\textbf{FULL}} \\
\cmidrule(lr){2-3}\cmidrule(lr){4-5}\cmidrule(lr){6-7}\cmidrule(lr){8-9}
& \textbf{F1$\uparrow$} & $\mathbf{IoU\uparrow}$ & \textbf{F1$\uparrow$} & $\mathbf{IoU\uparrow}$ & \textbf{F1$\uparrow$} & $\mathbf{IoU\uparrow}$ & \textbf{F1$\uparrow$} & $\mathbf{IoU\uparrow}$ \\
\midrule
4DMOS$^\ddagger$ \cite{mersch2022ral} & 25.9 & 18.5 & 8.4 & 6.1 & 0.6 & 0.4 & 24.4 & 16.7 \\
4DMOS \cite{mersch2022ral} & 47.3 & 32.1 & 22.7 & 15.4 & 8.3 & 5.6 & 31.8 & 21.6 \\
\bottomrule
\end{tabular}
\label{tab:evaluation_mos}
    \vspace{-2mm}
\end{table}

\begin{table}[t]
\centering
\caption{\textbf{3D Reconstruction Quality Results.} We report PSNR and SSIM for a NeRF and two 3D Gaussian Splatting methods.}
\vspace{-2mm}
\setlength{\tabcolsep}{11pt}
\scriptsize
\begin{tabular}{lccc}
\toprule
\textbf{Method} & \textbf{Representation} & \textbf{PSNR $\uparrow$} & \textbf{SSIM $\uparrow$} \\
\midrule
Dyn.\ Nerfacto \cite{nerfstudio} & NeRF & 26.2870 & 0.8653 \\
HUGS \cite{Zhou_2024_CVPR}           & 3DGS & 29.2675 & 0.8858               \\
OmniRe \cite{chen2025omnire}         & 3DGS &{33.8244} & {0.9515}              \\
\bottomrule
\end{tabular}
\label{tab:nerfgs}
\vspace{-2mm}
\end{table}

\subsection{End2End Driving}
Collectively, all tasks above aim at enabling end-to-end planning aligned with TruckDrive’s goal of safe, reliable and proactive operation. 
We train and evaluate UniAD~\cite{hu2023_uniad} as a recent E2E driving method, extending the ROI from $50$\,m to $250$\,m and replacing the original camera-only \cite{li2022bevformer} BEV backbone with a LiDAR-based architecture \cite{bai2022transfusionrobustlidarcamerafusion}, offering a first E2E benchmark for long-range highway driving. 
We evaluate UniAD on open-loop planning with standard L2 error, see results in table~\ref{tab:planning_uniad}.

\begin{table}[t]
\centering
\caption{\textbf{E2E Planning.} We train UniAD \cite{hu2023_uniad} on our long range setup and evaluate L2 error for all predicted time intervals.}
\vspace{-2mm}
\label{tab:planning_uniad}
\setlength{\tabcolsep}{7pt}
\resizebox{\linewidth}{!}{
\begin{tabular}{lccccc ccc}
\toprule
\multicolumn{1}{l}{\multirow{2}{*}{\textbf{Method}}} &
\multicolumn{7}{c}{\textbf{$L2$ (m)$\downarrow$}} \\
\cmidrule(lr){2-8}
& \textbf{1 step} & \textbf{2 step} & \textbf{3 step} & \textbf{4 step} & \textbf{5 step} & \textbf{6 step} & \textbf{Avg.} \\
\midrule
UniAD \cite{hu2023_uniad} & 0.57 & 1.13	& 1.71 & 2.30 & 2.88 & 3.42 & 2.00 \\
\bottomrule
\end{tabular}
}
\vspace{-2mm}
\end{table}

\section{Discussion}
\label{sec:discussion}
Our experiments confirm that across all tasks, existing model architectures designed for publicly available short-range data underperform when trained on TruckDrive’s long-range regime, with scores monotonically dropping with distance. Camera-only models exhibit the lowest performance, with average $57$\% lower mAP for $2$D object detection and up to $99$\% lower mAP for $3$D object detection (Far3D \cite{jiang2023far3dexpandinghorizonsurroundview}) in far (LR) distances. Architectures relying on camera, limited by compute constraints, necessitate $3\times$ downsampling of native $8$MP inputs, substantially degrading performance; for instance, Long Range stereo depth estimation exhibits an $8\times$ MAE increase (BridgeDepth \cite{guan2025bridgedepth}) due to reduced pixel disparities. LiDAR based and fusion-based architectures are aided in training by the additional long range $3$D representation, but struggle in sustaining the high dimensional complexity of the data and the large translation of objects. As existing methods largely rely on dense BEV representations, extending the maximum range forces either larger grids with fixed resolution, inducing a quadratic memory growth, or coarser cells with fixed grid dimension, degrading localization and association of both smaller objects and far-range instances, as shown Figure \ref{fig:qualitative_resuls}. As a result, $3$D multi-object tracking performs poorly (average $10$\% AMOTA), and we observe drops up to $83\%$ for moving-object segmentation (4DMOS \cite{mersch2022ral}) and up to $31\%$ for long-range $3$D object detection (BEVFusion \cite{liu2022bevfusion}). Finally, UniAD \cite{hu2023_uniad} requires extensive down-sampling across the entire architecture to allow the model to fit in the memory. The $250\times250$ meters ROI is encoded in a $200\times200$ BEV grid over the entire implementation, too coarse to encode useful driving information and not accurate enough to compute meaningful collision metric values. Overall, the model struggles to achieve low L2 planning error even for close future timestamps ($3$ step: $1.71$\,m), showcasing how urban-centric architectures fail to scale to long-range and high speed scenarios, highlighting the need for further research to unlock safe and reliable highway driving.

\section{Conclusion}
\label{sec:conclusion}
We introduce an autonomous driving dataset with $2$D annotations up to $1$\,km and $3$D annotations up to $400$\,m tailored for highway driving. While existing datasets focus on urban passenger car driving, the proposed TruckDrive dataset aims at opening up the research to highway driving where higher speed requires the ego agent to use different trajectories and maneuvers. We specifically focus on heavy-duty commercial trucks, which present an additional layer of complexity due the immense mass and break system lags extending the useful perception range from $80$\,m to $400$\,m.

Our evaluations on the dataset expose a persistent gap between state-of-the-art methods and the requirements of trucking highway autonomy. Hence, the dataset establishes a benchmark for range-aware, temporally grounded and computationally efficient driving methods that operate safely and reliably at high speed over long distances, and serves as a foundation for future research into driving methods tailored to the unique challenges of highway-scale autonomy, still far less explored than their urban counterpart.

\section{Acknowledgments}
Felix Heide was supported by an NSF CAREER Award (2047359), a Packard Foundation Fellowship, a Sloan Research Fellowship, a Sony Young Faculty Award, a Project X Innovation Award and a Amazon Science Research Award. Felix Heide is a co-founder of Algolux (now Torc Robotics), Head of AI at Torc Robotics, and a co-founder of Cephia AI. \\ 
We thank everyone at Torc Robotics for making TruckDrive possible, in particular the teams responsible for sensor integration, fleet operations, data engineering, and tooling for dataset generation and curation. 

{
    \small
    \bibliographystyle{ieeenat_fullname}
    \bibliography{main}
}


\end{document}